\documentclass{article}
\usepackage[utf8]{inputenc}
\usepackage{spconf,amsmath,graphicx}

\usepackage{amssymb}
\usepackage[usenames]{color}
\usepackage{makecell}
\usepackage{hyperref}
\usepackage{multirow}
\usepackage[table]{xcolor}
\usepackage{enumitem}
\usepackage{rotating}
\usepackage{array}
\usepackage{caption}
\usepackage{subcaption}
\usepackage{booktabs}
\usepackage{soul}
\usepackage{cuted}   
\usepackage{caption} 
\usepackage{enumitem} 

\title{3DPCNet: Pose Canonicalization for Robust Viewpoint-Invariant \\ 3D Kinematic Analysis from Monocular RGB cameras}
\name{Tharindu Ekanayake$^{\star}$, Constantino Álvarez Casado$^{\star}$$^{\dagger}$, and Miguel Bordallo López$^{\star}$}
\address{
$^{\star}$Center for Machine Vision and Signal Analysis (CMVS), University of Oulu, Finland \\
$^{\dagger}$Candour Ltd, Oulu, Finland \\
}
\begin{document}

\maketitle

\begin{strip}
\vspace{-5mm}
\centering
\includegraphics[width=\textwidth]{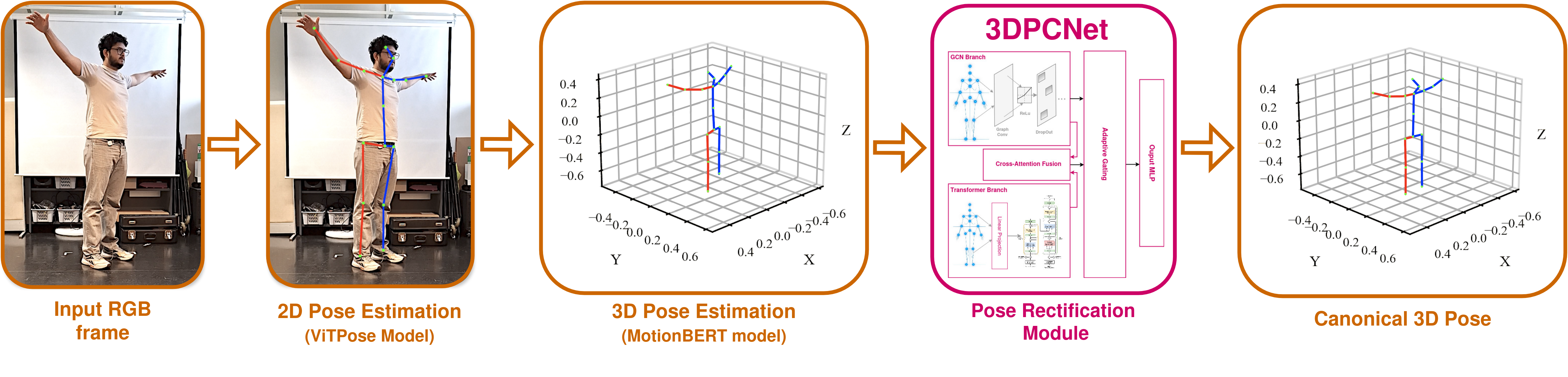}
\vspace{-9mm}
\captionof{figure}{Proposed 3DPCNet pipeline for view-invariant pose analysis. A monocular RGB frame yields a camera-centered 3D skeleton via 2D keypoint detection and 3D lifting. The 3DPCNet module then transforms this skeleton into a fixed, body-centered canonical frame by predicting a global rotation. This rectification removes camera viewpoint distortions, allowing for comparable kinematic measurements (e.g., wrist trajectories, joint angles) across different recordings. Code available at: \url{https://github.com/tharindu326/3DPCNet}}
\label{fig:3dpcnet_pipeline}
\end{strip}

\begin{abstract}

Monocular 3D pose estimators produce camera-centered skeletons, creating view-dependent kinematic signals that complicate comparative analysis in applications such as health and sports science. We present 3DPCNet, a compact, estimator-agnostic module that operates directly on 3D joint coordinates to rectify any input pose into a consistent, body-centered canonical frame. Its hybrid encoder fuses local skeletal features from a graph convolutional network with global context from a transformer via a gated cross-attention mechanism. From this representation, the model predicts a continuous 6D rotation that is mapped to an $SO(3)$ matrix to align the pose. We train the model in a self-supervised manner on the MM-Fi dataset using synthetically rotated poses, guided by a composite loss ensuring both accurate rotation and pose reconstruction. On the MM-Fi benchmark, 3DPCNet reduces the mean rotation error from over 20$^{\circ}$ to 3.4$^{\circ}$ and the Mean Per Joint Position Error from ~64 mm to 47 mm compared to a geometric baseline. Qualitative evaluations on the TotalCapture dataset further demonstrate that our method produces acceleration signals from video that show strong visual correspondence to ground-truth IMU sensor data, confirming that our module removes viewpoint variability to enable physically plausible motion analysis.

\end{abstract}

%
%
\section{Introduction}
\label{sec:intro}

Comparable kinematic measurements are essential in applications such as health monitoring \cite{jokinen2025understanding}, sports science \cite{pueo2017application,ortiz2022survey}, and virtual reality training \cite{li2025systematic}. For example, rehabilitation protocols track limb symmetry, and athletic coaching involves detailed analysis of joint trajectories \cite{ekanayake2025evaluating}. While monocular video is a practical source for full-body motion data, its output is fundamentally tied to the camera coordinate frame. Consequently, if the person or the camera moves, the same physical motion produces different 3D joint trajectories, which distorts kinematic signals like wrist paths or joint angle curves and complicates longitudinal analysis.

Although state-of-the-art 3D monocular pose estimators have improved localization accuracy \cite{martinez2017simple,pavllo20193d,Wang2019Robust,MotionBERT}, they do not solve this core problem, as the predicted skeleton remains camera-centered. This sensitivity of monocular pipelines to the viewpoint is well documented \cite{bhoi2019monoculardepthestimationsurvey,kiciroglu2020activemocapoptimizedviewpointselection}. Existing workarounds are insufficient for robust kinematic analysis. Rule-based normalizations that align the hips or shoulders are not robust to arbitrary rotations. Image-driven view adaptation methods require pixel access and camera calibration, which are unavailable in pose-only analysis pipelines. Although multi-view systems can reduce ambiguity, they also increase the cost and complexity of the setup.

This work addresses view-dependent variability by introducing \textit{3DPCNet}, a compact module that maps any estimated 3D pose, represented as joint coordinates $X \in \mathbb{R}^{J \times 3}$, to a fixed body-centered canonical frame. The system is agnostic to the upstream 3D pose estimator and is trained using only 3D pose data. The architecture uses a hybrid encoder with parallel graph convolutional and transformer branches, fused via a gated cross-attention mechanism. This encoder predicts a continuous 6D representation of the global rotation, which is mapped to a valid rotation matrix in $SO(3)$ using Gram-Schmidt orthonormalization. This matrix rectifies the input pose. An optional head can also predict a residual correction to fine-tune joint positions after the primary rotation. We evaluate our method on two benchmarks. Experiments on the MM-Fi dataset \cite{Yang2023mmfi} validate the accuracy of our canonical pose estimation. On the TotalCapture dataset \cite{Trumble2017BMVC}, we address a key motivation of our work by demonstrating that our rectification enables the extraction of acceleration signals from video that more closely resemble the ground truth from contact-based IMU sensors. Our primary contribution is a pose-only rectification module with a hybrid GCN-Transformer encoder that predicts a global 6D rotation and an optional residual correction to map any 3D skeleton into a canonical, body-centered frame. This module is trained using a self-supervised strategy with synthetically rotated poses, guided by a composite objective function.


    
    
    

%
%
\section{Related Work}
\label{sec:relatedwork}

Monocular 3D pose estimation methods provide the raw data for kinematic analysis, typically through direct image-to-3D regression or, more commonly, a two-stage approach that "lifts" 2D keypoint detections to 3D \cite{martinez2017simple}. State-of-the-art temporal models, including transformer- and diffusion-based architectures such as MotionBERT \cite{MotionBERT} and DDHPose \cite{ddhpose}, have significantly improved accuracy over earlier methods \cite{pavllo20193d}. However, a fundamental property of these pipelines is that they produce skeletons in a camera-centered coordinate frame. This means the numerical representation of a motion changes with the camera's orientation, which is a major limitation for applications requiring consistent measurements \cite{Mercadal2024Exercise}.

Several approaches have been developed to address this viewpoint sensitivity, particularly in the context of skeleton-based action recognition. These methods often learn to transform input skeletons into a common, virtual viewpoint to achieve higher classification accuracy \cite{zhang2019view,liu2017enhanced}. While effective for recognition, such transformations are not optimized to preserve the metric fidelity of the pose, which is critical for precise kinematic analysis. Within the 3D pose estimation field itself, some works adapt the 2D pose to a more favorable view before the lifting stage \cite{Liang2020Adaptive}. Other research, such as CanonPose \cite{wandt2021canonpose}, focuses on disentangling the canonical body pose from the camera rotation in a self-supervised manner, even this process is often coupled with the image-based feature extraction stage. The technical feasibility of learning rotations directly from pose data is supported by work on continuous 6D rotation representations \cite{zhou2019continuity} and models that jointly estimate joint orientations and positions \cite{fisch2021orientation}.


Despite these developments, most prior work either focuses on improving reconstruction or recognition accuracy, or tightly couples view adaptation to image features. Less attention has been paid to creating a modular 3D-to-3D rectification module that operates independently of the upstream pose estimator. Our work addresses this gap by learning the global rotation that maps any camera-centered skeleton to a fixed frontal canonical frame, using losses that align the rotation on $SO(3)$ and preserve the underlying pose structure. This design provides a flexible post-processing stage to stabilize joint-level trajectories for any monocular pipeline.

\section{Proposed Methodology}
\label{sec:methods}

This section details the proposed 3D Pose Canonicalization Network (3DPCNet). Our objective is to learn a mapping function that transforms an input 3D skeleton $X \in \mathbb{R}^{J \times 3}$, where $J$ is the number of joints, from an arbitrary camera-centered view into a consistent, predefined canonical orientation. We frame this as a rotation estimation problem, where the model predicts the global rotation that aligns the input pose with the canonical frame.

\subsection{Pose Canonicalization Architecture}
\label{sec:architecture}

\begin{figure}[ht!]
 \begin{center}
  \includegraphics*[width=0.49\textwidth]{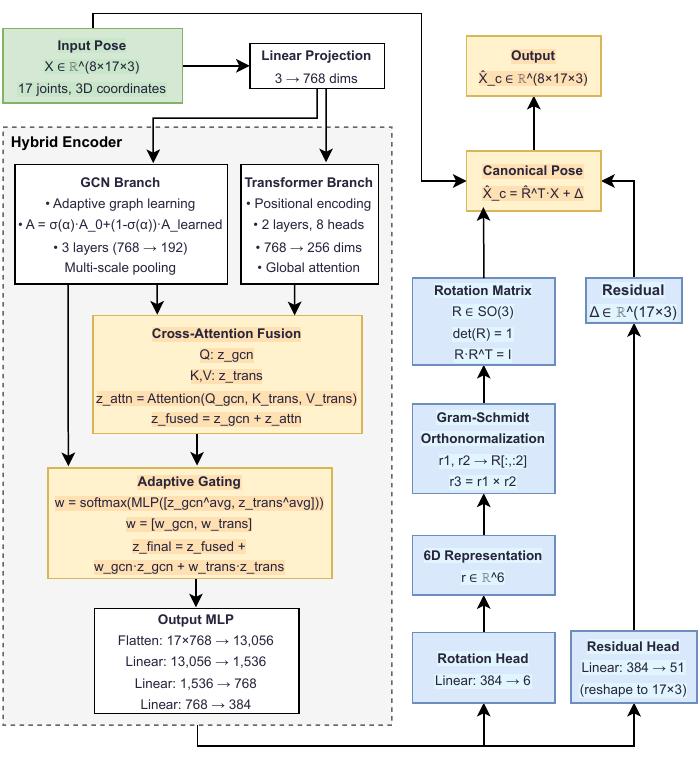}
 \end{center}
 \vspace{-3mm}
 \caption{The architecture of 3DPCNet. An input pose $X \in \mathbb{R}^{17 \times 3}$ is processed by parallel GCN and Transformer branches. The resulting features are fused via cross-attention and a learned gate. Two heads then predict a rotation (as a 6D vector) and an optional residual correction $\Delta$. The final canonical pose is computed as $\hat{X}_c = \hat{R}^\top X + \Delta$ and supervised through a composite loss function.}
 \label{fig:model_architecture}
\end{figure}

The core of our method is a network that takes a centered 3D pose with $J=17$ joints and predicts the canonical representation. As illustrated in Figure~\ref{fig:model_architecture}, the architecture has two primary stages: a hybrid encoder for feature extraction and two prediction heads for generating the final output.

\vspace{-4mm}
\subsubsection{Hybrid Encoder for Pose Feature Extraction}
The encoder, $f_\theta$, processes the input pose through two parallel streams designed to capture both local and global spatial relationships between joints:

\vspace{2mm}
\noindent\textbf{Graph Convolutional Branch:} The first stream models the explicit anatomical structure of the human body using a Graph Convolutional Network (GCN). The GCN operates on an adaptive adjacency matrix $A$, which is a learned interpolation between the fixed anatomical skeleton graph, $A_0$, and a learnable graph topology, $A_{\text{learned}}$. This is formulated as \(A = \sigma(\alpha) A_0 + (1-\sigma(\alpha)) A_{\text{learned}}\), where $A_{\text{learned}}$ is a learnable symmetric matrix and $\sigma(\cdot)$ is the sigmoid function applied to a learnable parameter $\alpha$. This adaptive structure allows the model to learn non-anatomical yet task-relevant connections between joints.

\vspace{2mm}
\noindent\textbf{Transformer Branch:} The second stream uses a Transformer encoder to model long-range dependencies that are not captured by the local connectivity of the GCN. The input joint coordinates are first projected into an embedding space and combined with learnable positional encodings. The resulting sequence is processed by several multi-head self-attention layers, allowing the model to capture global context and relationships between distant joints, such as a hand and the contralateral foot.

\vspace{2mm}
\noindent\textbf{Feature Fusion:} The fusion mechanism combines both representations through cross-attention, where GCN features query Transformer representations to produce attended features. An adaptive gating network learns task-specific weights $[w_{\text{gcn}}, w_{\text{trans}}]$ to balance the contributions in the final fused feature vector $z$, defined as \( z = z_{\text{fused}} + w_{\text{gcn}} \cdot z_{\text{gcn}} + w_{\text{trans}} \cdot z_{\text{trans}}\). This adaptive fusion allows the model to leverage local skeletal constraints when beneficial while utilizing global context for complex pose configurations.

\subsubsection{Rotation and Residual Prediction Heads}
From the fused feature vector $z$, two separate heads predict the components needed to generate the canonical pose:

\vspace{1mm}
\noindent\textbf{Rotation Prediction.}
The primary head predicts a continuous 6D representation for the rotation, following the approach in \cite{zhou2019continuity}. This 6D vector is then converted into a valid 3x3 rotation matrix, $\hat{R} \in SO(3)$, using the Gram-Schmidt orthonormalization process. This parameterization avoids discontinuities associated with other rotation representations like Euler angles or quaternions, leading to more stable training.

\vspace{1mm}
\noindent\textbf{Canonical Pose Computation.}
The estimated rotation matrix $\hat{R}$ represents the transformation from the canonical frame to the input camera view. Therefore, its transpose, $\hat{R}^\top$, is applied to the input pose $X$ to reverse the rotation. An optional second head predicts a residual correction, $\Delta \in \mathbb{R}^{J \times 3}$, to fine-tune joint positions. The final canonical pose, $\hat{X}_c$, is computed as \(\hat{X}_c = \hat{R}^\top X + \Delta\). This formulation separates the rigid global rotation from small, non-rigid pose adjustments.

\subsection{Training and Optimization}
\label{sec:training}

The model is trained end-to-end using a loss function that integrates multiple objectives to ensure both accurate rotation estimation and structurally consistent pose reconstruction.

\vspace{-2mm}
\subsubsection{Proposed Loss Function}
The total loss, $\mathcal{L}_{\text{total}}$, is a weighted sum of several components. The primary objective is the \textit{pose reconstruction los}s, $\mathcal{L}_{\text{pose}}$, which is the mean squared error between the predicted canonical pose $\hat{X}_c$ and the ground-truth canonical pose $X_c$. To directly supervise the rotation, we use the \textit{geodesic distance} on the $SO(3)$ manifold, which measures the angle of the relative rotation between the predicted matrix $\hat{R}$ and the ground-truth matrix $R_{\text{GT}}$. It is calculated as:
\begin{equation}
\mathcal{L}_{\text{rot}} = \arccos\left(\frac{\text{tr}(\hat{R}R_{\text{GT}}^\top) - 1}{2}\right)
\label{eq:geodesic_loss}
\end{equation}
where $\text{tr}(\cdot)$ is the trace of the matrix. A \textit{cycle consistency loss}, $\mathcal{L}_{\text{cyc}}$, ensures that applying the predicted forward rotation $\hat{R}$ to the predicted canonical pose $\hat{X}_c$ reconstructs the original input pose $X$. This is formulated as \( \mathcal{L}_{\text{cyc}} = \|\hat{R}\hat{X}_c - X\|_2^2 \). Additional terms regularize the training. A \textit{perceptual loss}, $\mathcal{L}_{\text{perc}}$, penalizes deviations in bone lengths and joint angles. Regularization losses, $\mathcal{L}_{\text{reg}}$, include an L2 penalty on the residual term $\Delta$ to discourage large corrections, a term to maintain a meaningful topology in the learned GCN graph, and an attention diversity term for the Transformer heads. The complete objective function is given by
\begin{equation}
\mathcal{L}_{\text{total}} = w_p \mathcal{L}_{\text{pose}} + w_r \mathcal{L}_{\text{rot}} + w_c \mathcal{L}_{\text{cyc}} + w_{\text{perc}} \mathcal{L}_{\text{perc}} + w_{\text{reg}} \mathcal{L}_{\text{reg}},
\label{eq:total_loss}
\end{equation}
where the $w$ terms are scalar weights that balance the contribution of each component.

\subsubsection{Implementation Details}
The hybrid encoder has a hidden dimension of 256 and an output dimension of 384. It contains 3 GCN layers and 2 Transformer layers with 8 attention heads. We train the model for 80 epochs using the AdamW optimizer with a learning rate of $5 \times 10^{-4}$ and a cosine annealing schedule. The batch size is 1024. The primary loss weights are $w_p=1.0$ and $w_r=1.0$, with smaller weights for the auxiliary terms ($w_c=0.25$, $w_{\text{perc}}=0.15$).

\subsection{Geometric Baseline}
\label{sec:geometric_baseline}
For comparison, we implement a deterministic geometric canonicalization method. This approach does not require learned parameters and is based on anatomical landmarks. It first defines a plane using the shoulder and hip joints and rotates the pose to align this plane's normal vector with a fixed axis (e.g., the Z-axis). A second rotation is then applied to align the vector between the shoulders parallel to the X-axis. This two-step process deterministically produces a front-facing canonical pose and serves as a rule-based baseline in our experiments.

%
%
\section{Experimental Evaluation}
\label{sec:evaluation}

We designed two experiments to evaluate 3DPCNet. The first experiment assesses the core performance of pose canonicalization on a dedicated benchmark. The second evaluates the practical utility of the canonicalized poses by measuring the stability and physical plausibility of derived kinematic signals with real-world sensor data.

\subsection{Benchmark Datasets and Preprocessing}
\label{sec:datasets}

\subsubsection{MM-Fi for Canonicalization Training and Evaluation}
We use the MM-Fi dataset \cite{Yang2023mmfi} as the primary source for 3D skeleton data. It contains recordings of 40 subjects in 27 activities across four environments. For our task, we generate input-target pairs from the provided ground-truth 3D poses ($X \in \mathbb{R}^{17 \times 3}$). First, each pose is centered at the pelvis and mapped to a consistent canonical coordinate system where the subject faces the negative X-axis. This remapped, centered pose serves as the ground-truth target, $X_{\text{canonical}}$. To simulate varied camera viewpoints, we generate a corresponding input, $X_{\text{input}}$, by applying a random rotation $R \in SO(3)$ to the canonical pose, given by \( X_{\text{input}} = R \cdot X_{\text{canonical}}\). The rotation matrix $R$ is constructed from sampled yaw, pitch, and roll angles designed to emulate realistic camera placements (e.g., mostly frontal views with some side and back views), while avoiding extreme or physically implausible orientations. This process creates physically valid rotated poses that preserve bone lengths and joint angles. We follow the official MM-Fi evaluation protocols: S1 (cross-sequence), S2 (cross-subject), and S3 (cross-environment).

\subsubsection{TotalCapture for Kinematic Signal Analysis}
\label{ssec:totalcapture_dataset}

\begin{figure}[ht!]
\centering
\includegraphics[width=0.49\textwidth]{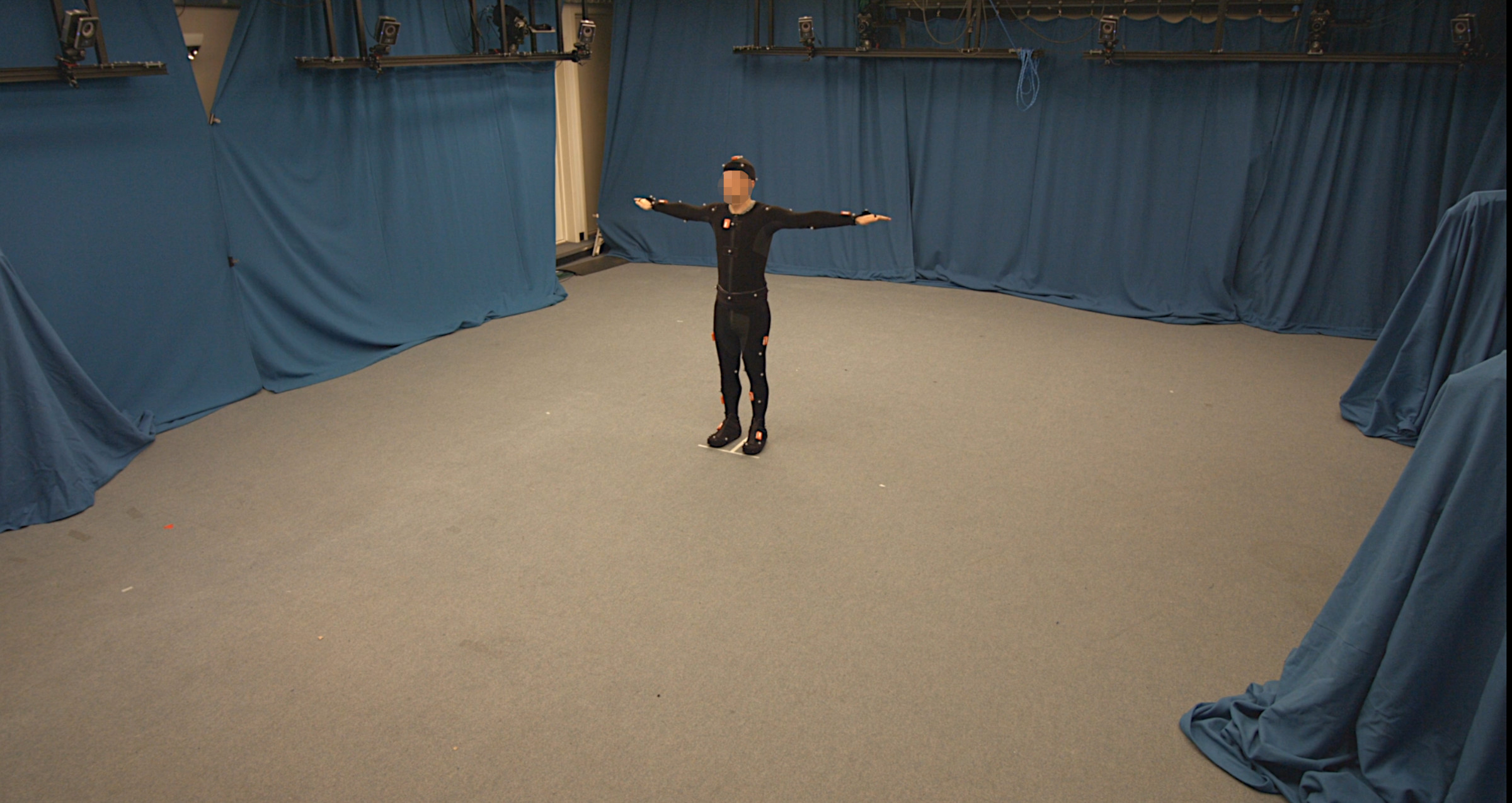} 
\caption{TotalCapture example frame \cite{Trumble2017BMVC}: Camera~8, Subject~1, \emph{Acting 2}. Multi-view RGB is time-aligned with Vicon and 13 IMUs, enabling cross-modal evaluation of pose-derived kinematics.}
\label{fig:tc_cam8_rgb}
\end{figure}

We use the TotalCapture (CVSSP3D) dataset \cite{Trumble2017BMVC}\footnote{\url{https://cvssp.org/data/cvssp3d/}} to assess cross–modal agreement between pose-derived and inertial kinematics. TotalCapture provides fully synchronized, gen-locked \emph{eight-view} RGB video together with Vicon motion capture (21 joints) and \emph{13} Xsens IMUs across \(\sim\!1.9\)M frames spanning five subjects and activities including ROM, Walking, Freestyle, and Acting. Camera and IMU calibrations are supplied in the “Surrey” format, which we use to rotate per-sensor accelerations from the local IMU frame into the global world frame and then remove gravity. Vicon joint positions are converted from inches to meters and remapped to our 17-joint layout for consistency. We follow the official split: training on subjects \(\{1,2,3\}\) with ROM\{1,2,3\}, Walking\{1,3\}, Freestyle\{1,2\}, Acting\{1,2\}; testing on subjects \(\{1,2,3,4,5\}\) with Walking~2, Freestyle~3, Acting~3 (unseen sequences, mix of seen/unseen subjects). Importantly, 3DPCNet is neither trained nor fine-tuned on TotalCapture. The dataset is used solely to evaluate whether canonicalized 3D motion derived from monocular pipelines matches IMU dynamics. Figure~\ref{fig:tc_cam8_rgb} shows an example RGB frame (Camera~8, Subject~1, \emph{Acting 2}) used for qualitative context. 

\subsection{Evaluation Metrics}
\label{sec:metrics}
We evaluate our method canonicalization performance using three standard metrics. We measure the absolute positional accuracy using the Mean Per Joint Position Error (MPJPE), which is the average Euclidean distance between predicted and ground-truth joints. To assess the structural fidelity of the pose shape independent of its global orientation, we use the Procrustes-Aligned MPJPE (PA-MPJPE). Finally, we directly evaluate the primary task of orientation alignment by computing the Rotation Error, defined as the geodesic distance in degrees between the predicted and ground-truth rotation matrices, as formulated in Equation~\eqref{eq:geodesic_loss}.

%
%
\subsection{Results on Pose Canonicalization}
\label{sec:canon_results}

We evaluated the estimation of canonical pose in MM-Fi using the S2 (cross subject) and S3 (cross environment) protocols using MPJPE, PA-MPJPE and rotation error. As summarized in Table~\ref{tab:canon_results}, 3DPCNet reduces positional error from \(63\text{–}65\) mm to \(46\text{–}49\) mm (\(\approx\!27\%\) reduction) and crucially reduces global orientation error from \(20\text{–}22^\circ\) to \(3\text{–}4^\circ\) (5–6\(\times\) improvement). Performance is stable across domains: moving from S2 to S3 changes MPJPE by only \(\sim\)1~mm and rotation error by \(\lesssim 1^\circ\), indicating that the learned rectifier generalizes across subjects and environments without access to images or calibration.

The behavior of PA-MPJPE clarifies the role of our residual refinement. The GEOMETRIC baseline performs a rigid reorientation and therefore attains PA-MPJPE \(=0\) by construction (shape is preserved and Procrustes removes the remaining rotation/scale). In contrast, 3DPCNet yields \(37\text{–}38\)~mm PA-MPJPE, reflecting \emph{small, learned non-rigid adjustments} that denoise upstream 3D estimates (e.g., slight limb length or joint placement inconsistencies) while still producing a consistent canonical frame. In practice, these corrections improve downstream kinematic readouts (Sec.~\ref{sec:motion_results}) by reducing view-induced distortions and stabilizing joint trajectories.

\begin{table}[ht!]
\renewcommand{\arraystretch}{1.2}
\setlength{\tabcolsep}{3.0pt}
\centering
\caption{Canonicalization on MM-Fi. MPJPE and PA-MPJPE in mm, rotation error in degrees.}
\vspace{-2mm}
\label{tab:canon_results}
\begin{tabular}{|l l | c c c|}
\hline
\textbf{Dataset} & \textbf{Model} & \textbf{MPJPE} & \textbf{PA-MPJPE} & \textbf{Rot. Err.} \\
\hline
\multirow{3}{*}{MM-Fi S2} 
& GEOMETRIC   & 62.85 & 0.00  & 20.64 \\
& 3DPCNetS2   & 47.57  & 37.49 & 3.58  \\
& 3DPCNetS3   & 46.24  & 36.94 & 3.43  \\
\hline
\multirow{3}{*}{MM-Fi S3} 
& GEOMETRIC   & 64.58 & 0.00  & 21.56 \\
& 3DPCNetS2   & 48.83  & 38.22 & 4.29  \\
& 3DPCNetS3   & 47.71  & 37.61 & 4.24  \\
\hline
\end{tabular}
\end{table}

\begin{figure}[ht!]
\begin{center}
\includegraphics[width=0.48\textwidth]{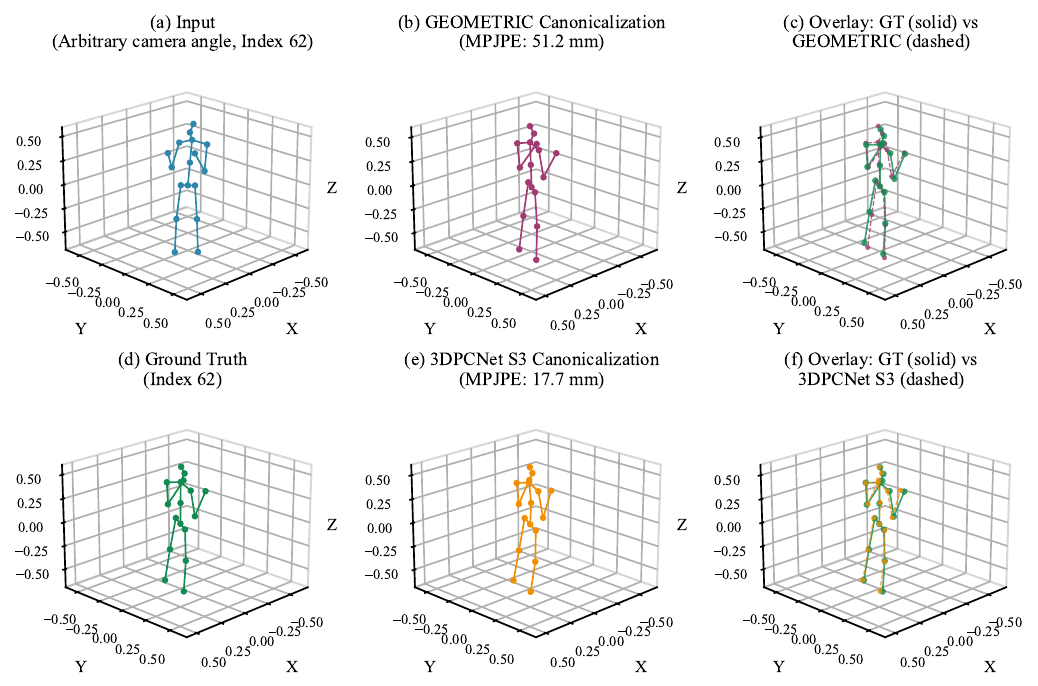}
\end{center}
\vspace{-3mm}
\caption{Qualitative comparison on an S3 test sample. Top row shows input, GEOMETRIC output, and overlay with ground truth. Bottom row shows ground truth, 3DPCNet output, and overlay.}
\label{fig:canonicalization_comparison}
\end{figure}

Figure~\ref{fig:canonicalization_comparison} illustrates these effects in a representative sample of S3. The GEOMETRIC output roughly aligns the torso, but leaves a noticeable misalignment in the distal joints when overlaid with ground truth (Fig.~\ref{fig:canonicalization_comparison}c), whereas 3DPCNet S3 produces a rotation that nearly coincides with the canonical target across the torso and limbs (Fig.~\ref{fig:canonicalization_comparison}f). The per-sample MPJPEs shown in the figure (GEOMETRIC 51.2~mm vs.\ 3DPCNet~17.7~mm) are consistent with the average trend in Table~\ref{tab:canon_results}. Overall, the results demonstrate that learning a pose-only rectification, with an explicit \(SO(3)\) rotation and a light residual, yields a compact, estimator-agnostic module that materially reduces viewpoint bias and delivers canonical skeletons suitable for joint-level analysis.

\subsection{Results on Kinematic Trajectory Stability}
\label{sec:motion_results1}

\begin{figure}[ht!]
\vspace{-3mm}
\begin{center}
\includegraphics[width=0.49\textwidth]{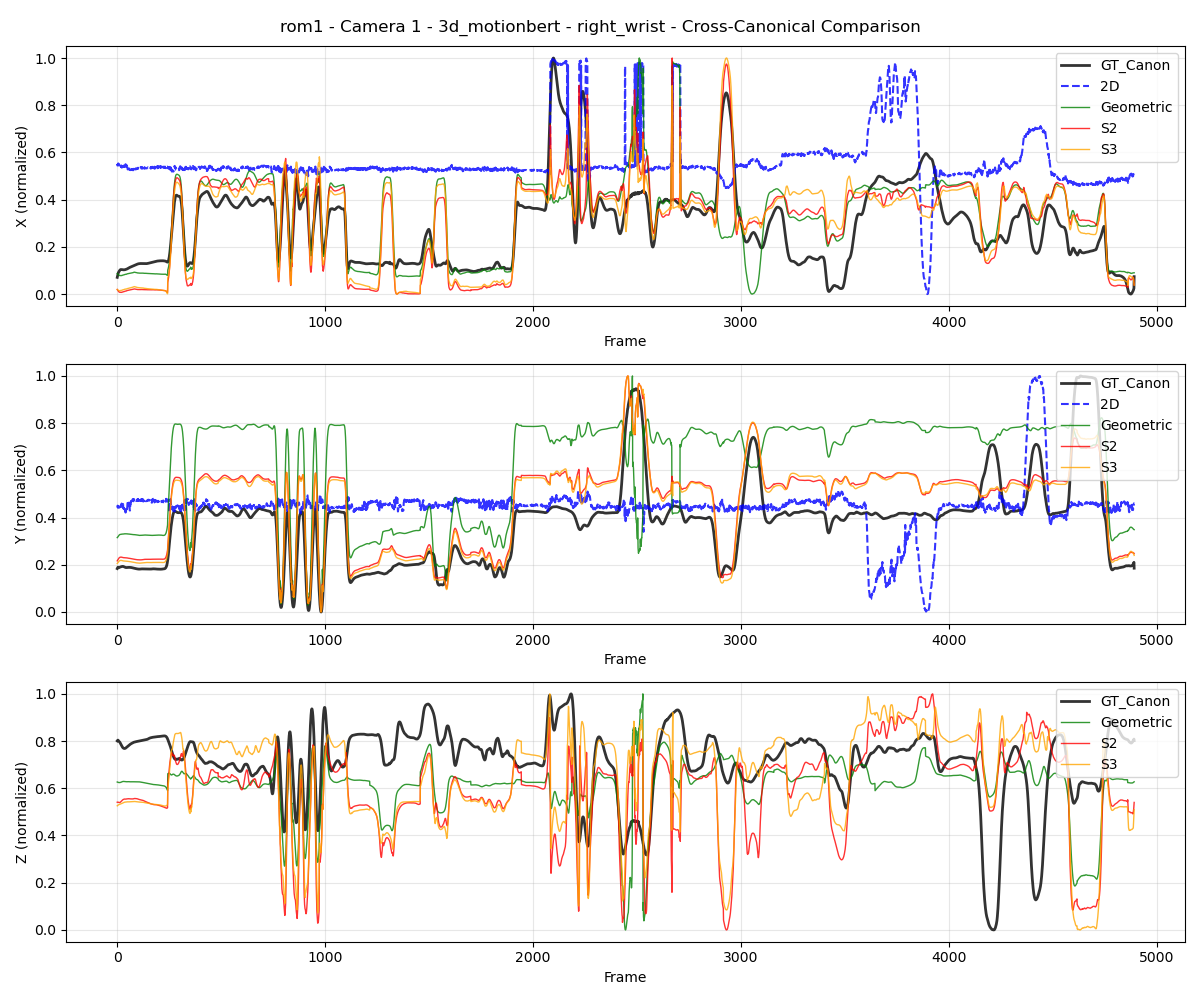} 
\end{center}
\vspace{-6mm}
\caption{Normalized right wrist trajectories (X, Y, Z axes) during a ROM exercise. The trajectories from our 3DPCNet models (S2, red; S3, orange) track the ground-truth canonical signal (black) more closely than both the unstable 2D signal (blue, dashed) and the geometric baseline (green).}
\label{fig:cross_canonical_wrist}
\end{figure}

To evaluate the practical impact of our method, we assessed its ability to produce stable and accurate kinematic trajectories from poses generated by a state-of-the-art estimator (MotionBERT). We compare the normalized 3D wrist trajectories produced by different canonicalization methods against a ground-truth canonical trajectory obtained from a motion capture system. Figure~\ref{fig:cross_canonical_wrist} provides a qualitative comparison for a representative Range of Motion (ROM) exercise. The results clearly demonstrate the necessity of 3D canonicalization. The trajectory derived from raw 2D keypoints (blue, dashed) is highly erratic and deviates significantly from the ground truth (black), confirming its unsuitability for direct analysis. In contrast, all 3D canonicalized trajectories are far more stable and closely follow the reference path.

Crucially, the trajectories rectified by our 3DPCNet models (red and orange) show a superior alignment with the ground truth compared to the geometric baseline (green). As seen across all three axes, but most clearly in the Z-axis plot, our learned models more faithfully capture the high-frequency details and overall dynamics of the true motion. The geometric method produces a reasonable approximation but tends to smooth out or slightly miss the more complex parts of the trajectory. This visual evidence suggests that 3DPCNet not only enforces a view-invariant reference frame but also produces higher-fidelity kinematic signals, which is essential for detailed motion analysis in clinical and performance settings.

\subsection{Results on Cross-Modal Kinematic Fidelity}
\label{sec:motion_results}

\begin{figure}[ht!]
\vspace{-3mm}
\begin{center}
\includegraphics[width=0.49\textwidth]{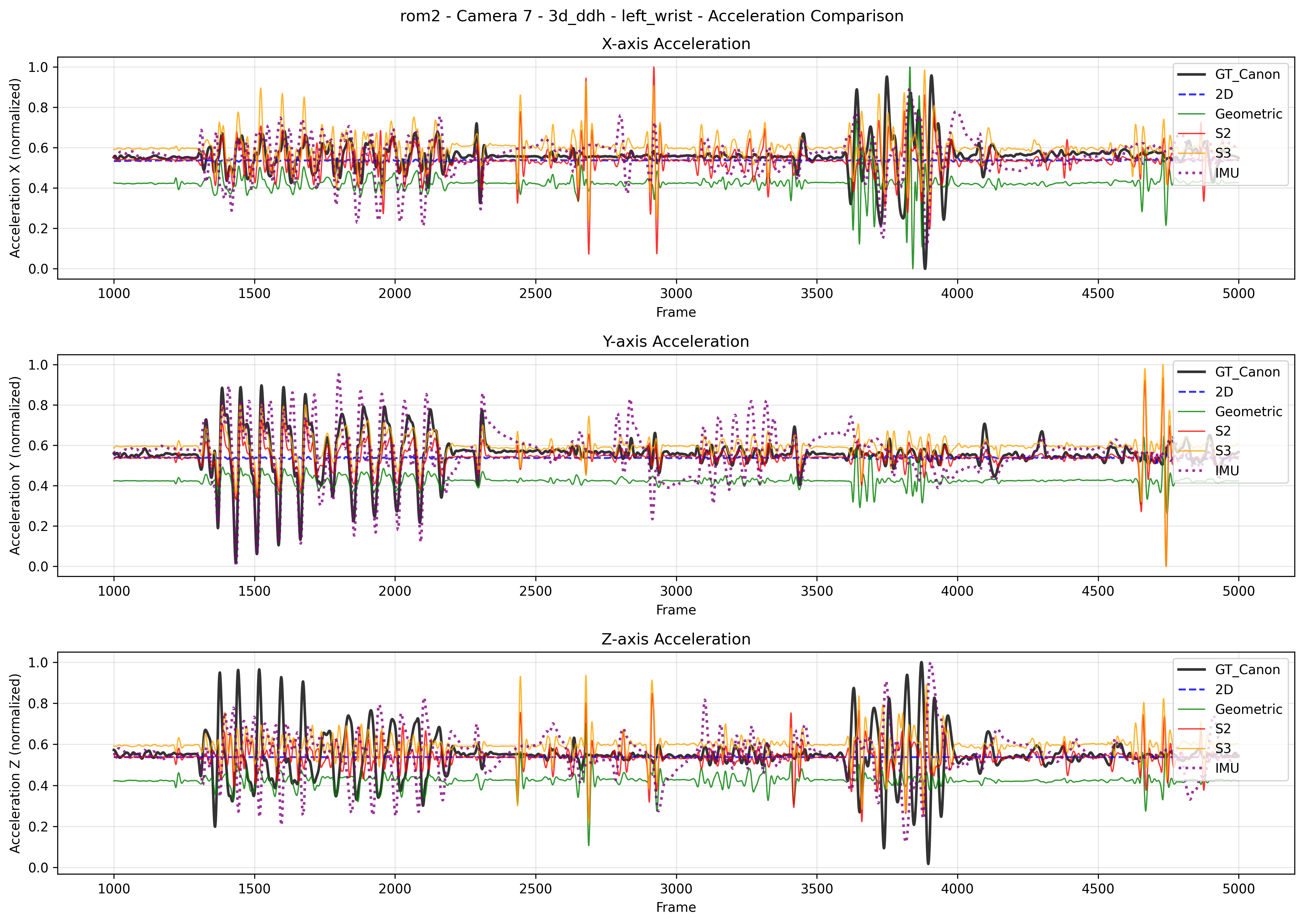} 
\end{center}
\vspace{-6mm}
\caption{Normalized wrist acceleration signals from a challenging lateral viewpoint (TotalCapture, Camera 7). The signals from our 3DPCNet models (S2, red; S3, orange) show a strong visual correspondence with the ground-truth IMU signal (purple, dashed), outperforming the geometric baseline (green).}
\label{fig:acceleration_comparison}
\end{figure}

To evaluate if 3DPCNet produces physically plausible kinematics, we derived wrist acceleration signals from DDH-estimated poses \cite{ddhpose} on the TotalCapture dataset \cite{Trumble2017BMVC} and compared them against a synchronized IMU ground truth. To allow a valid comparison, the raw IMU data was transformed from local sensor coordinates to the global frame with gravity removed. All signals in Figure~\ref{fig:acceleration_comparison} are normalized. Crucially, our model, trained only on MM-Fi, was tested on a challenging lateral view (Camera 7) from this new dataset without any fine-tuning to evaluate its generalization. 

The results in Figure~\ref{fig:acceleration_comparison} presents a qualitative comparison of the normalized acceleration signals. The results demonstrate a strong visual correspondence between the signals derived from our canonicalized poses (3DPCNet-S2 and S3, red and orange lines) and the ground-truth IMU signal (purple, dashed). Our models successfully capture the primary dynamic patterns of the motion, including the timing, shape, and relative intensity of the acceleration peaks. In contrast, the signal from the **Geometric** baseline (green) shows more noticeable deviations and appears less consistent with the IMU data. This close alignment is particularly significant given the challenging camera view, which can introduce errors into the initial 3D pose. Our learned model appears to be better equipped to handle these imperfections than the rigid rule-based geometric approach. This visual evidence suggests that 3DPCNet not only enforces a view-invariant frame but also preserves the underlying physical dynamics of the movement, producing a high-fidelity kinematic signal that more closely approximates data from contact-based sensors.

%
%
\section{Conclusion}
\label{sec:conclusion}


This work tackled a persistent obstacle for monocular human-motion analysis: the lack of a stable, body-centered reference frame that makes joint-level signals comparable across cameras, sessions, and subjects. We introduced 3DPCNet, a pose-only, estimator-agnostic rectification module that predicts a global rotation in \(SO(3)\) (plus a light residual) to map any camera-centered skeleton into a predefined canonical frame. In MM-Fi, 3DPCNet reduced the rotation error from \(20\text{–}22^\circ\) to \(3\text{–}4^\circ\) and MPJPE by \(\approx 27\%\), with negligible degradation from cross-subject evaluation (S2) to cross-environment (S3). Beyond numeric gains, qualitative IMU comparisons on TotalCapture \cite{Trumble2017BMVC} show that canonicalized motion derived from off-the-shelf pose estimators tracks inertial dynamics more faithfully than a geometric baseline, especially under challenging lateral viewpoints. Together, these results indicate that a dedicated pose-only canonicalization stage can substantially diminish viewpoint-induced variability and produce kinematic trajectories better suited for downstream analysis in rehabilitation, sports and VR training.

\textbf{Insights and significance.} (i) Separating rigid orientation from small non-rigid corrections is effective: the learned rotation delivers view invariance, while the residual acts as a denoiser that compensates for upstream pose biases without requiring images or calibration. (ii) The hybrid GCN–Transformer encoder and cross-attentive fusion exploit both anatomical structure and long-range joint dependencies, which proves crucial for accurate global orientation prediction. (iii) Stability across S2/S3 splits and qualitative agreement with IMU signals suggest that canonicalization learned from pose-only supervision can generalize beyond the training domain and benefit sensor-fusion scenarios.

\textbf{Limitations.} Our residual head can introduce minor non-rigid deviations (reflected by non-zero PA-MPJPE), which may be undesirable when strict rigidity is required. The method does not recover absolute scale; applications needing metric distances must obtain scale from upstream estimators, anthropometric priors, or additional sensors. Performance still depends on the quality of the input 3D skeleton, where severe occlusions or extreme viewpoints that corrupt the upstream pose will propagate errors. Finally, our training relies on pose-space augmentation (sampled yaw/pitch/roll) rather than real multi-view imagery, which may under-represent rare poses and camera geometries.

\textbf{Outlook.} Future work will (i) constrain or explicitly control the residual toward \(\mathrm{SE}(3)\)-rigid behavior when desired (e.g., with stronger bone-length/joint-angle priors or differentiable Procrustes constraints), (ii) extend the model temporally to exploit video dynamics and produce uncertainty-aware canonical trajectories, (iii) integrate scale estimation and gravity alignment for absolute kinematics, and (iv) validate clinical and performance metrics end-to-end (e.g., ROM, asymmetry indices) in real deployments. We believe plug-and-play canonicalization is a practical path to make monocular pipelines deliver view-robust, quantitatively comparable kinematics at scale.

\section*{Acknowledgment}
The research was supported by the Business Finland WISEC project (Grant 3630/31/2024), the University of Oulu and the Research Council of Finland (former Academy of Finland) 6G Flagship Programme (Grant Number: $346208$), Profi5 HiDyn programme (326291), and Profi7 Hybrid intelligence program (352788). The authors acknowledge the CSC-IT Center for Science, Finland, for computational resources. The small transformer diagram inside the 3DPCNet block in Figure \ref{fig:3dpcnet_pipeline}, created by \textit{dvgodoy} and distributed via the repository \url{https://github.com/dvgodoy/dl-visuals} under the CC BY 4.0 license.

\bibliographystyle{IEEEbib}
\bibliography{references}

\begin{thebibliography}{10}

\bibitem{jokinen2025understanding}
Ella Jokinen, Ella P{\"a}{\"a}kk{\"o}nen, Constantino~{\'A}lvarez Casado, Aino~K Rantala, Terhi Ruuska-Loewald, Jouni~JK Jaakkola, and Miguel~Bordallo L{\'o}pez,
\newblock ``Understanding the feasibility of computer vision in diagnosing respiratory infections in pediatric emergency rooms,''
\newblock {\em medRxiv}, pp. 2025--04, 2025.

\bibitem{pueo2017application}
Basilio Pueo and Jose~Manuel Jimenez-Olmedo,
\newblock ``Application of motion capture technology for sport performance analysis (el uso de la tecnolog{\'\i}a de captura de movimiento para el an{\'a}lisis del rendimiento deportivo),''
\newblock {\em Retos}, vol. 32, pp. 241--247, 2017.

\bibitem{ortiz2022survey}
Vanessa~E Ortiz-Padilla, Mauricio~A Ram{\'\i}rez-Moreno, Gerardo Presb{\'\i}tero-Espinosa, Ricardo~A Ram{\'\i}rez-Mendoza, and Jorge de~J Lozoya-Santos,
\newblock ``Survey on video-based biomechanics and biometry tools for fracture and injury assessment in sports,''
\newblock {\em Applied Sciences}, vol. 12, no. 8, pp. 3981, 2022.

\bibitem{li2025systematic}
Xiaohui Li, Dongfang Fan, Junjie Feng, Yu~Lei, Chao Cheng, and Xiangnan Li,
\newblock ``Systematic review of motion capture in virtual reality: Enhancing the precision of sports training,''
\newblock {\em Journal of Ambient Intelligence and Smart Environments}, vol. 17, no. 1, pp. 5--27, 2025.

\bibitem{ekanayake2025evaluating}
Tharindu Ekanayake, Constantino {\'A}lvarez~Casado, Nhi Nguyen, Marta Sobocinski, Sari Pramila-Savukoski, Xiaoting Wu, Kristina Mikkonen, and Miguel Bordallo~L{\'o}pez,
\newblock ``Evaluating the accuracy and reliability of camera-based physiological and motion signal extraction techniques in virtual reality training environments,''
\newblock in {\em Scandinavian Conference on Image Analysis}. Springer, 2025, pp. 442--456.

\bibitem{martinez2017simple}
Julieta Martinez, Rayat Hossain, Javier Romero, and James~J Little,
\newblock ``A simple yet effective baseline for 3d human pose estimation,''
\newblock in {\em Proceedings of the IEEE international conference on computer vision}, 2017, pp. 2640--2649.

\bibitem{pavllo20193d}
Dario Pavllo, Christoph Feichtenhofer, David Grangier, and Michael Auli,
\newblock ``3d human pose estimation in video with temporal convolutions and semi-supervised training,''
\newblock in {\em Proceedings of the IEEE/CVF conference on computer vision and pattern recognition}, 2019, pp. 7753--7762.

\bibitem{Wang2019Robust}
Chunyu Wang, Yizhou Wang, Zhouchen Lin, and Alan~L. Yuille,
\newblock ``Robust 3d {Human} {Pose} {Estimation} from {Single} {Images} or {Video} {Sequences},''
\newblock {\em IEEE Transactions on Pattern Analysis and Machine Intelligence}, vol. 41, no. 5, pp. 1227--1241, may 1 2019.

\bibitem{MotionBERT}
Wentao Zhu, Xiaoxuan Ma, Zhaoyang Liu, Libin Liu, Wayne Wu, and Yizhou Wang,
\newblock ``Motionbert: A unified perspective on learning human motion representations,'' 2023.

\bibitem{bhoi2019monoculardepthestimationsurvey}
Amlaan Bhoi,
\newblock ``Monocular depth estimation: A survey,'' 2019.

\bibitem{kiciroglu2020activemocapoptimizedviewpointselection}
Sena Kiciroglu, Helge Rhodin, Sudipta~N. Sinha, Mathieu Salzmann, and Pascal Fua,
\newblock ``Activemocap: Optimized viewpoint selection for active human motion capture,'' 2020.

\bibitem{Yang2023mmfi}
Jianfei Yang, He~Huang, Yunjiao Zhou, Xinyan Chen, Yuecong Xu, Shenghai Yuan, Han Zou, Chris~Xiaoxuan Lu, and Lihua Xie,
\newblock ``Mm-fi: multi-modal non-intrusive 4d human dataset for versatile wireless sensing,''
\newblock in {\em Proceedings of the 37th International Conference on Neural Information Processing Systems}, Red Hook, NY, USA, 2023, NIPS '23, Curran Associates Inc.

\bibitem{Trumble2017BMVC}
Matt Trumble, Andrew Gilbert, Charles Malleson, Adrian Hilton, and John Collomosse,
\newblock ``Total capture: 3d human pose estimation fusing video and inertial sensors,''
\newblock in {\em 2017 British Machine Vision Conference (BMVC)}, 2017.

\bibitem{ddhpose}
Qingyuan Cai, Xuecai Hu, Saihui Hou, Li~Yao, and Yongzhen Huang,
\newblock ``Disentangled diffusion-based 3d human pose estimation with hierarchical spatial and temporal denoiser,'' 2025.

\bibitem{Mercadal2024Exercise}
Clara Mercadal-Baudart, Chao-Jung Liu, Garreth Farrell, Molly Boyne, Jorge Gonz{\' a}lez~Escribano, Aljosa Smolic, and Ciaran Simms,
\newblock ``Exercise quantification from single camera view markerless 3d pose estimation,''
\newblock {\em Heliyon}, vol. 10, no. 6, pp. e27596, 3 2024.

\bibitem{zhang2019view}
Pengfei Zhang, Cuiling Lan, Junliang Xing, Wenjun Zeng, Jianru Xue, and Nanning Zheng,
\newblock ``View adaptive neural networks for high performance skeleton-based human action recognition,''
\newblock {\em IEEE transactions on pattern analysis and machine intelligence}, vol. 41, no. 8, pp. 1963--1978, 2019.

\bibitem{liu2017enhanced}
Mengyuan Liu, Hong Liu, and Chen Chen,
\newblock ``Enhanced skeleton visualization for view invariant human action recognition,''
\newblock {\em Pattern Recognition}, vol. 68, pp. 346--362, 2017.

\bibitem{Liang2020Adaptive}
Guoqiang Liang, Xiangping Zhong, Lingyan Ran, and Yanning Zhang,
\newblock ``An {Adaptive} {Viewpoint} {Transformation} {Network} for 3d {Human} {Pose} {Estimation},''
\newblock {\em IEEE Access}, vol. 8, pp. 143076--143084, 2020.

\bibitem{wandt2021canonpose}
Bastian Wandt, Marco Rudolph, Petrissa Zell, Helge Rhodin, and Bodo Rosenhahn,
\newblock ``Canonpose: Self-supervised monocular 3d human pose estimation in the wild,''
\newblock in {\em Proceedings of the IEEE/CVF conference on computer vision and pattern recognition}, 2021, pp. 13294--13304.

\bibitem{zhou2019continuity}
Yi~Zhou, Connelly Barnes, Jingwan Lu, Jimei Yang, and Hao Li,
\newblock ``On the continuity of rotation representations in neural networks,''
\newblock in {\em Proceedings of the IEEE/CVF conference on computer vision and pattern recognition}, 2019, pp. 5745--5753.

\bibitem{fisch2021orientation}
Martin Fisch and Ronald Clark,
\newblock ``Orientation keypoints for 6d human pose estimation,''
\newblock {\em IEEE Transactions on Pattern Analysis and Machine Intelligence}, vol. 44, no. 12, pp. 10145--10158, 2021.

\end{thebibliography}

\end{document}